\documentclass{article}

\usepackage{microtype}
\usepackage{graphicx}
\usepackage{subcaption}
\usepackage{multirow}
\usepackage{tabularx}
\usepackage{booktabs} 

\usepackage{hyperref}

\usepackage[accepted]{icml2026}

\makeatletter
\renewcommand{\ICML@appearing}{\textit{Accepted at the AI for Science workshop (ICML 2026).}
Copyright 2026 by the author(s).}
\makeatother

\usepackage{amsmath}
\usepackage{amssymb}
\usepackage{mathtools}
\usepackage{amsthm}

\usepackage[capitalize,noabbrev]{cleveref}

\theoremstyle{plain}

\theoremstyle{definition}

\theoremstyle{remark}

\usepackage[disable,textsize=tiny]{todonotes}

\icmltitlerunning{Geo-Expert: Towards Expert-Level Geological Reasoning via
Parameter-Efficient Fine-Tuning}

\begin{document}

\twocolumn[
  \icmltitle{Geo-Expert: Towards Expert-Level Geological Reasoning via Parameter-Efficient Fine-Tuning}



  \icmlsetsymbol{equal}{*}

  \begin{icmlauthorlist}
    \icmlauthor{Chenyou Guo}{equal,ouc}
    \icmlauthor{Zongqi Liu}{equal,ouc}
    \icmlauthor{Yizhou Zhang}{equal,ouc}
    \icmlauthor{Zhaorui Jiang}{equal,pku,monash}
    \icmlauthor{Ze Liu}{ouc}
  \end{icmlauthorlist}

  \icmlaffiliation{ouc}{Ocean University of China}
  \icmlaffiliation{pku}{Peking University}
  \icmlaffiliation{monash}{Monash University}
  \icmlcorrespondingauthor{Zhaorui Jiang}{zrjiang25@stu.pku.edu.cn}
  \icmlcorrespondingauthor{Ze Liu}{liuze@ouc.edu.cn}

  \icmlkeywords{Large Language Models, Geological Reasoning, Parameter-Efficient Fine-Tuning, Instruction Synthesis, AI for Science, ICML}

  \vskip 0.3in
]



\printAffiliationsAndNotice{\icmlEqualContribution}

\begin{abstract}
  While general-purpose Large Language Models (LLMs) applied to Geology often hallucinate when reasoning about subsurface structures and deep-time evolution, current AI in Earth sciences predominantly targets surface remote sensing and GIS. To bridge this gap, we introduce Geo-Expert, a family of parameter-efficient geological LLMs fine-tuned on a custom-curated, high-quality instruction dataset processed with a reproducible synthesis workflow and geology-specific prompts and filters. We investigate the impact of model scaling and architecture by fine-tuning three base models: Qwen3-8B, Qwen3-32B, and Gemma-3-27B, with Low-Rank Adaptation (LoRA) method. Our extensive evaluation on a novel domain-specific benchmark, Geo-Eval, reveals that a domain-aligned 8B model can outperform open-weight 70B generalists and proprietary GPT-4o on specialized geological reasoning, while a 32B variant approaches frontier reasoning models. The optimized 8B model further offers a favorable parameter-efficiency profile for deployment. This work provides a reproducible recipe for democratizing scientific LLMs and establishes a baseline for geological artificial intelligence.
\end{abstract}

\section{Introduction}

Geology plays a vital role in underground resource exploration, geologic hazard mitigation, and understanding planetary evolution. Unlike general knowledge domains, geological reasoning requires interpreting complex spatio-temporal relationships and a large amount of professional data. However, LLMs often struggle with the specialized logic required in geology. General-purpose models frequently exhibit severe hallucinations when asked to perform a rigorous deduction on stratigraphy or structural geology, limiting their utility in geological research and industrial applications.

Current large models in the vertical field of geography predominantly focus on Geography Knowledge Agent and Earth Observation, demonstrating proficiency in general geospatial intelligence and utilizing computer vision to analyze imagery for surface monitoring. They effectively handle tasks related to surface geography, environmental monitoring, and human-geography interactions. However, a critical gap remains between general \textit{Geography} and specialized \textit{Geology}. While existing models can retrieve facts about Earth's surface, they often lack the deep-domain reasoning required to understand the solid Earth such as interpreting subsurface stratigraphic sequences, reconstructing deep-time tectonic evolution, or analyzing complex petrogenetic processes.

To bridge this gap, we introduce Geo-Expert, a suite of parameter-efficient geological LLMs. We construct a high-quality, domain-specific instruction dataset derived from five canonical geology textbooks. Using a document-to-instruction synthesis workflow, we synthesize structured instruction-response pairs with geology-specific prompts and filters to instill rigorous domain knowledge. We investigate model scaling and architecture effects by fine-tuning three base models: Qwen3-8B, Qwen3-32B, and Gemma-3-27B, employing Low-Rank Adaptation (LoRA) to improve training efficiency.

In sum, our main contributions are as follows:
\begin{itemize}
  \item \textbf{Dataset Construction:} We present a reproducible pipeline for converting unstructured geological textbooks into high-quality instruction-tuning datasets.
  \item \textbf{Model Scaling Analysis:} We provide a comparative analysis of model architectures (Qwen vs. Gemma) and parameter scales (8B vs. 27B/32B), showing that smaller models can be efficient deployment candidates while larger models remain helpful for complex structural reasoning.
  \item \textbf{Geo-Expert Benchmark:} We establish Geo-Eval, a benchmark for geological reasoning, demonstrating that our fine-tuned models substantially outperform general-purpose baselines in specialized tasks.
\end{itemize}

\begin{figure*}[t]
\vskip 0.2in
\begin{center}
\centerline{\includegraphics[width=\textwidth]{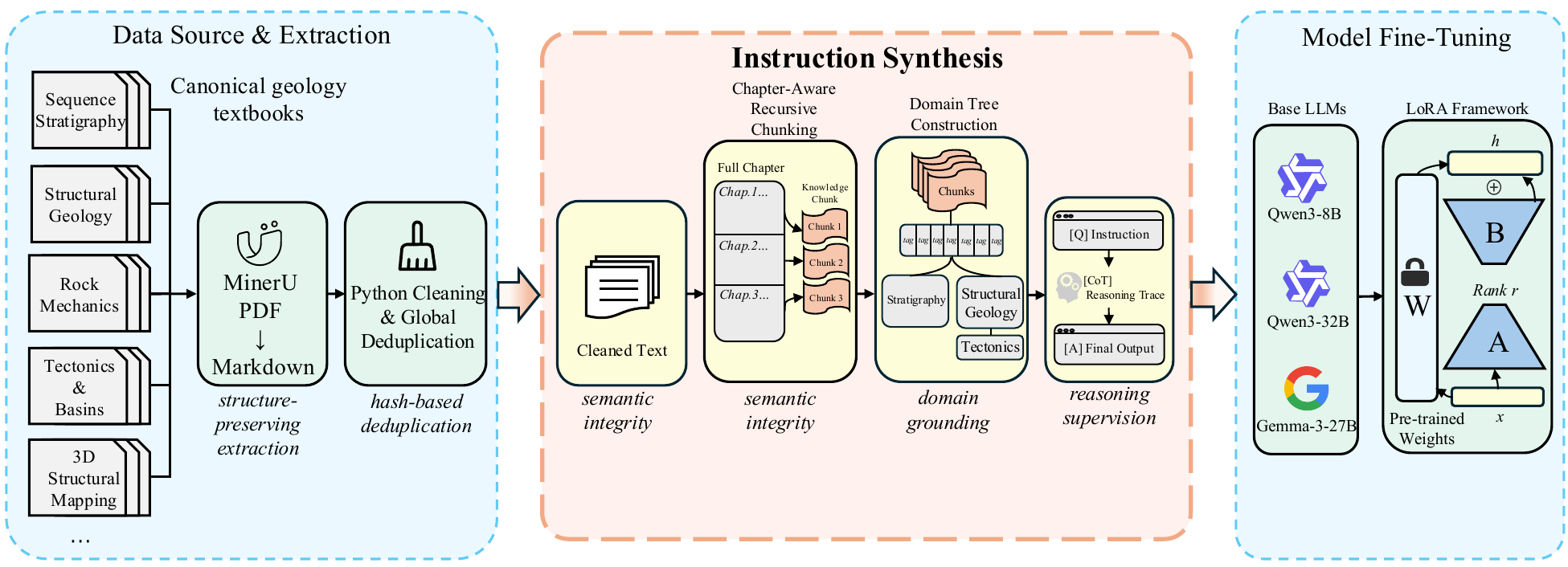}}
\caption{\textbf{Overview of the Geo-Expert framework.} The pipeline consists of three main stages: (1) extracting and sanitizing text from canonical geology textbooks; (2) synthesizing high-quality, CoT-enhanced instruction pairs via a domain-structured generation pipeline; (3) applying parameter-efficient fine-tuning on base LLMs.}
\label{fig:pipeline}
\end{center}
\vskip -0.2in
\end{figure*}

\section{Model Framework and Benchmark Construction}

\subsection{Data Sources}

Geological reasoning is fundamentally grounded in established physical principles and deep-time evolutionary models, necessitating data sources of the highest scientific rigor. To construct a high-fidelity instruction dataset, we identified five canonical textbooks as the data source. These sources were strategically selected to cover the full spectrum of solid earth science, balancing theoretical depth, cognitive difficulty, and practical application:

\begin{itemize}
\item \textit{Principles of Sequence Stratigraphy I \& II} (Catuneanu) \citep{catuneanu2006principles} --- This text serves as the standard for understanding temporal frameworks in sedimentation and basin analysis. It is renowned for its highly abstract conceptual models, challenging the model's deep-time evolutionary reasoning.
\item \textit{Structural Geology} \& \textit{Advanced Structural Geology} \citep{fossen2016structural} --- These volumes provide the theoretical foundation for rock mechanics, stress analysis, and rheology. They feature a steep learning curve with complex mathematical derivations, establishing a rigorous physical baseline.
\item \textit{Tectonics and Sedimentation: Implications for Petroleum Systems} \citep{gao2012tectonics} --- A critical resource bridging the gap between plate tectonics, basin formation, and hydrocarbon exploration. It provides complex, interdisciplinary macro-scale scenarios.
\item \textit{Structural Analysis and Synthesis} \citep{rowland2020structural} --- It focuses on practical application, including 3D geometric interpretation and structural mapping techniques. This volume strictly evaluates the model's spatial imagination and geometric deduction capabilities.
\end{itemize}

\subsection{Dataset Construction}

Transforming static textbook content into a dynamic instruction-tuning dataset involves a pipeline designed to maximize data fidelity and reduce noise. We implemented a multi-stage process: digitization, heuristic cleaning, and instruction synthesis.

\paragraph{Text Extraction and Sanitization.}
We first used MinerU \citep{mineru2024}, a high-precision PDF-to-Markdown tool, to extract textual content while preserving the structure of the original textbooks. Following the extraction, we developed a custom Python-based cleaning module to process the raw Markdown. This pipeline first segments the text into discrete blocks based on paragraph lines containing metadata keywords like page numbers and copy notices. Furthermore, to prevent redundancy caused by layout overlaps, we implemented a global deduplication mechanism using a hash-based method to filter out identical text blocks across the entire corpus.

\paragraph{Instruction Synthesis.}

To transform the cleaned Markdown segments into a high-quality, geoscientific instruction-tuning corpus, we implement a lightweight workflow for document chunking, question generation, answer generation, and dataset export. Our domain-specific contribution is the geology source curation, prompt design, filtering rules, and downstream training/evaluation protocol. The synthesis process consists of three key stages:
\begin{enumerate}
\item \textbf{Chapter-Aware Recursive Chunking:} Unlike simple fixed-length splitting, the framework utilizes a semantic segmentation algorithm based on the Markdown header structure. This ensures that each text chunk retains complete semantic context within its hierarchical scope, preventing the fragmentation of complex geological descriptions.
\item \textbf{Domain-Structured Question Generation:} An LLM first constructs a hierarchical domain tree (extracting outlines and concepts) to bind specific tags to text segments. Questions are then dynamically generated based on these domain tags and character density, minimizing redundancy and ensuring broad coverage of geological concepts.
\item \textbf{Chain-of-Thought (CoT) Answer Construction:} To capture the rigorous logic required for geological reasoning, the system employs reasoning-oriented teacher models, primarily DeepSeek-R1 with source-grounded prompts, to generate candidate answers. Each generated item is constrained to cite or paraphrase the provided chunk rather than external memory, and malformed or source-inconsistent outputs are filtered before training. This process produces not only the final answer but also intermediate reasoning traces, improving supervision for multi-step deduction while keeping the answer aligned with the source material.
\end{enumerate}
This automated pipeline enabled us to efficiently scale the dataset to 11,518 high-quality instruction pairs while preserving the authoritative accuracy of the original canonical sources.

\subsection{Model Fine-Tuning}

We performed instruction tuning using the ms-swift framework \citep{msswift2024}, applying distinct hyperparameter strategies tailored to the scale of each model. We targeted all linear layers (all-linear) for Low-Rank Adaptation (LoRA) \citep{hu2021lora}.

For the compact Qwen3-8B model \citep{qwen2024qwen25}, we adopted a standard efficiency configuration with LoRA rank $r$=32, $a$=32, and a learning rate of 2e-5 using FP16 precision. This lightweight setup, with gradient checkpointing enabled, allowed for rapid iteration on a single NVIDIA RTX 5090.

In contrast, to effectively adapt the deeper architectures of Gemma-3-27B \citep{gemma2024gemma} and Qwen3-32B, we scaled up the adaptation capacity, increasing the LoRA rank to $r$=64 and alpha to $a$=128. To mitigate the high memory footprint and ensure numerical stability during training, we employed BF16 (bfloat16) precision and activated gradient checkpointing with a gradient accumulation step of 4. This optimized configuration enabled the fine-tuning of these 27B+ parameter models on a 4$\times$RTX 5090 setup, demonstrating a feasible recipe for fine-tuning 27B+ models on consumer-grade GPUs.

\subsection{Benchmark Construction: Geo-Eval}

Standard benchmarks in Earth sciences predominantly focus on remote sensing tasks or general geography trivia, failing to capture the rigorous deductive logic required for solid earth geology. To bridge this evaluation gap and comprehensively evaluate the true reasoning capabilities of LLMs in the vertical domain of geology, we constructed Geo-Eval, a novel domain-specific benchmark consisting of high-fidelity, open-ended reasoning questions. Unlike traditional static datasets manually extracted from exams, Geo-Eval was constructed through an automated adversarial mining pipeline coupled with expert human-in-the-loop verification. The construction process proceeded in four distinct phases:
\begin{enumerate}
\item \textbf{Large-Scale Generative Extraction:} We initially deployed DeepSeek-R1 \citep{deepseek2025r1} to perform a comprehensive semantic analysis of the source geology textbooks. Through this process, we extracted a foundational pool of 2,591 complex geological questions and synthesized their corresponding preliminary reference answers.
\item \textbf{Comparative Inference:} To evaluate the baseline domain capability, both our fine-tuned Qwen3-8B-Geo and DeepSeek-R1 were tasked with generating independent responses to the entire pool of 2,591 questions.
\item \textbf{Difficulty-Aware Evaluation Paradigm (LLM-as-a-Judge):} To move beyond static, undifferentiated testing, we introduce a \textit{difficulty-aware evaluation paradigm} designed to identify the true cognitive frontier of geological reasoning. We utilized GLM-4.5 \citep{glm2024glm4} as an impartial evaluator to blindly score both models' responses on a 10-point scale. Rather than evaluating on a random distribution, we systematically isolated a subset of 387 ``hard boundary'' questions where the score differential between our Qwen3-8B-geo and the frontier generalist DeepSeek-R1 was marginal (score difference $\le$ 4). This dynamic calibration of difficulty yields boundary samples where a strong generalist no longer has a decisive margin over the domain-aligned model. Consequently, this design shifts the focus from basic factual recall to a stricter, boundary-aware stress test of expert-level geological reasoning.
\item \textbf{Expert Vetting and Ground Truth Finalization:} To improve scientific validity, the isolated 387 candidate questions were submitted to geology professors for manual screening, refinement, and calibration. The final test set uses expert-written reference answers rather than model-generated answers.
\end{enumerate}
The resulting finalized dataset serves as a highly targeted, expert-verified ground truth. By deliberately focusing on these high-complexity boundary problems, Geo-Eval provides a robust metric for assessing whether a model has truly internalized deep geological logic, moving far beyond superficial pattern matching.

\subsection{Selection Bias and Benchmark Scope}
We acknowledge that the boundary mining process uses Qwen3-8B-Geo as one comparator, which can introduce selection bias because the initial candidate pool is partly shaped by this model's strengths and weaknesses. Our mitigations reduce, but do not eliminate, this coupling: (i) the 387 candidate questions were independently re-vetted, refined, and, where necessary, rewritten by geology professors, and the reference answers used for scoring are expert-authored ground truth, not model outputs; (ii) the final evaluation employs GPT-4o as a reference-guided judge against this expert-verified ground truth, decoupling scoring from the mining process; (iii) two of our three reported models (Qwen3-32B-Geo and Gemma-3-27B-Geo) were not used in the boundary-mining loop, yet exhibit consistent gains. We therefore interpret Geo-Eval as a hard, boundary-aware stress test rather than a fully independent estimate of performance over the entire geology question distribution. Constructing a larger benchmark from an entirely model-independent sampling process remains important future work.

\begin{table*}[t]
\centering
\caption{Performance comparison of various models on the Geo-Eval Benchmark. The best results in each column are highlighted in bold, and the second-best are underlined. The $\Delta$ column indicates the absolute performance gain achieved by our Geo-Expert fine-tuning framework compared to the respective base model. $^{\ast\ast\ast}$ indicates statistical significance ($p = 3.7121 \times 10^{-106} < 0.001$) compared to the corresponding base models via paired t-test; this statistic should be interpreted within the boundary-mined benchmark rather than as an i.i.d. estimate over all geology questions.}
\label{tab:main_benchmark}
\begin{tabular}{lcccccc}
\toprule
\textbf{Model} & \textbf{Size} & \textbf{Concept} & \textbf{Process} & \textbf{Engineering} & \textbf{Average} & $\Delta$ \\
\midrule
\multicolumn{7}{c}{\textit{Proprietary SOTA}} \\
\midrule
GPT-5.4 (Frontier)  & -   & \textbf{7.35} & \textbf{7.10} & \textbf{7.00} & \textbf{7.15} & - \\
DeepSeek-V3.2-Instruct & -   & \underline{6.80} & 6.75 & 6.67 & 6.74 & - \\
GPT-4o                 & -   & 6.10 & 5.90 & 5.80 & 5.93 & - \\
\midrule
\multicolumn{7}{c}{\textit{Open-weight Generalists}} \\
\midrule
Gemma-3-27B-IT         & 27B & 5.30 & 5.10 & 5.08 & 5.16 & - \\
Qwen3-32B              & 32B & 5.20 & 4.90 & 4.90 & 5.00 & - \\
Qwen3-8B               & 8B  & 4.80 & 4.68 & 4.41 & 4.63 & - \\
GLM-4-9B               & 9B  & 4.70 & 4.50 & 4.51 & 4.57 & - \\
Llama-3.1-70B-Instruct & 70B & 4.30 & 4.10 & 3.96 & 4.12 & - \\
\midrule
\multicolumn{7}{c}{\textbf{Geo-Expert (Ours)}} \\
\midrule
\textbf{Qwen3-32B-geo}   & 32B & 6.78 & \underline{6.79} & \underline{6.90} & \underline{6.82}$^{\ast\ast\ast}$ & +1.82 \\
\textbf{Gemma-3-27B-geo} & 27B & 6.70 & 6.60 & 6.47 & 6.59$^{\ast\ast\ast}$ & +1.43 \\
\textbf{Qwen3-8B-geo}    & 8B  & 6.10 & 6.27 & 6.44 & 6.27$^{\ast\ast\ast}$ & +1.64 \\
\bottomrule
\end{tabular}
\end{table*}

\section{Benchmark and Evaluation: Geo-Eval}

\subsection{Benchmark Content}
To avoid one-dimensional evaluation, we rigorously partition the Geo-Eval test dataset into three cognitive levels:

\begin{itemize}
\item \textbf{Concept:} Focuses on the identification and definition of core terminology, such as mineralogy, petrology, and stratigraphy.
\item \textbf{Process:} Focuses on the logical deduction of dynamic geological mechanisms, including tectonic movements and diagenesis.
\item \textbf{Engineering:} Focuses on the comprehensive application analysis of geological conditions in practical scenarios, such as engineering site selection and geological hazard prevention.

\end{itemize}

Because Geo-Eval contains 387 expert-vetted questions, it is intended for targeted stress testing rather than exhaustive subdiscipline coverage. The three-way partition provides a coarse view of reasoning type, but the current size limits finer-grained claims within each category.

\subsection{Benchmark Baselines}
After applying Geo-Eval to different models, we adopt the widely recognized LLM-as-a-Judge paradigm, utilizing GPT-4o as a reference-guided LLM evaluator comparing against expert-verified ground truth. We impose strict domain-fact-checking constraints within the prompt to generate a quantitative score ranging from 0 to 10.
We select 11 highly representative models for a comprehensive comparison, categorized into three groups: proprietary state-of-the-art models (e.g., GPT-5.4, DeepSeek-V3.2), open-weight generalist models of varying parameter scales (e.g., Llama-3.1-70B, Qwen3-32B), and our domain-specific models fine-tuned via the proposed Geo-Expert framework.
We acknowledge that LLM-as-a-Judge evaluation can introduce evaluator-specific bias. The reference-guided prompt is designed to reduce free-form preference effects by anchoring scores to expert answers, but it is not a substitute for a full human reliability study. A multi-judge or human-adjudicated subset is an important next step for strengthening Geo-Eval.

\subsection{Main Results and Insights}
The experimental results demonstrate the advantages of the Geo-Expert framework on the boundary-aware Geo-Eval setting. We analyze the results from three core dimensions:
\begin{enumerate}
\item \textbf{Remarkable Performance Breakthrough}
As shown in Table 1, general open-weight base models struggle with high-depth geological questions. However, after fine-tuning with the Geo-Expert framework, all models achieve a substantial improvement in performance. For instance, the average score of Qwen3-32B rises from 5.00 to 6.82, and Gemma-3-27B rises from 5.16 to 6.59. Notably, in the ``Engineering'' dimension, which most directly tests applied deductive reasoning, the score of Qwen3-8B rises from 4.41 to 6.44 after fine-tuning. This suggests that, in specialized STEM branches, relying solely on general knowledge injected during pre-training can be insufficient. Furthermore, a paired t-test on the 387 samples shows that the performance gains of the Geo-Expert models are statistically significant within this benchmark ($p = 3.7121 \times 10^{-106} < 0.001$), although the non-random boundary-mined construction means that the statistic should be interpreted as evidence of consistent per-question gains on Geo-Eval rather than as a population-level estimate.
\item \textbf{Extreme Parameter Efficiency}
One of the most striking findings of this study is the parameter efficiency exhibited by the fine-tuned models. The Qwen3-8B-geo model, with 8B parameters, outperforms Llama-3.1-70B-Instruct across all three dimensions (overall score 6.27 vs. 4.12), and also exceeds the proprietary GPT-4o baseline (5.93) on this benchmark. This result suggests that domain alignment can be more valuable than parameter count in narrow expert settings. Conversely, a compact model with refined domain supervision may be a practical alternative to larger generalist models in resource-constrained geological scenarios.
\item \textbf{Competitive Frontier Comparison}
Our best-performing model, Qwen3-32B-geo, achieves a score of 6.82, exhibiting balance across the Concept (6.78), Process (6.79), and Engineering (6.90) dimensions. This performance exceeds GPT-4o in our evaluation and remains below the strongest proprietary baseline, GPT-5.4 (7.15), by 0.33 points on average. Considering its smaller parameter scale and local deployability, Geo-Expert models provide a practical cost-performance tradeoff for academic and engineering use.
\end{enumerate}

\subsection{Hardware Efficiency}
To make the deployment claim more concrete, Table \ref{tab:efficiency} summarizes training efficiency from our logs and a simple score-per-parameter proxy. The 8B model obtains the highest Geo-Eval score per billion parameters and can be trained on a single consumer GPU. We did not conduct a controlled inference-latency benchmark across all proprietary and open models because several baselines were accessed through APIs, but the local hardware profile indicates why the 8B model is the most practical deployment target.

\begin{table}[t]
\centering
\caption{Training and hardware-efficiency summary for Geo-Expert models. Runtime and memory are taken from the ms-swift training logs. Score/B is the Geo-Eval average divided by total model parameters in billions.}
\label{tab:efficiency}
\scriptsize
\setlength{\tabcolsep}{2pt}
\begin{tabular}{@{}lcccc@{}}
\toprule
\textbf{Model} & \textbf{Train.} & \textbf{GB} & \textbf{Hours} & \textbf{Score/B} \\
\midrule
Qwen3-8B-geo & 87M & 21.8 & 2.23 & 0.757 \\
Gemma-3-27B-geo & 454M & 87.5 & 4.97 & 0.236 \\
Qwen3-32B-geo & 537M & 106.9 & 6.03 & 0.205 \\
\bottomrule
\end{tabular}
\end{table}

\subsection{Qualitative Analysis}
\label{sec:qualitative_analysis}
To intuitively demonstrate the effectiveness of the Geo-Expert framework in mitigating catastrophic domain-shifting errors, we present a qualitative comparison between GPT-4o and our fine-tuned model (Qwen3-8B-geo) on a highly specialized structural geology question, as shown in Table \ref{tab:qualitative_case}.

As illustrated in Table \ref{tab:qualitative_case}, the base model, lacking deep geological corpus alignment, exhibits a severe cross-domain hallucination. It incorrectly interprets the polysemous term ``wedge'' as a mechanical or civil engineering component, generating completely irrelevant advice such as using carbon fiber cloth for concrete reinforcement. This literal interpretation leads to a complete failure in answering the geological prompt.

In stark contrast, the Geo-Expert fine-tuned model successfully anchors the semantic context to structural geology. It accurately explains the mechanical responses (stress concentration, fault reactivation) and correctly identifies the complex tectonic accommodation mechanisms (thrust sliding, lateral material flow). This striking contrast (0 vs. 9) highlights a critical vulnerability of generalist LLMs: when encountering polysemous terms across disciplines, they tend to default to high-frequency general concepts. The Geo-Expert framework effectively recalibrates the model's prior distribution, ensuring precise semantic anchoring and rigorous factual reasoning in the target scientific domain.

\begin{table}[t]
\centering
\caption{Qualitative comparison of model outputs. The base model suffers from cross-domain hallucination, whereas Geo-Expert accurately anchors to geology.}
\label{tab:qualitative_case}
\small
\begin{tabularx}{\linewidth}{@{}X@{}}
\toprule
\textbf{Prompt:} \\
What phenomena does the local thickening of a wedge cause, and how is this deformation accommodated? \\
\midrule
\textbf{GPT-4o:} \hfill \textbf{[Score: 0/10]} \\
\textit{Cross-domain Hallucination:} \\
...Local thickening may alter the structural stress distribution, leading to \textbf{concrete cracking} or gear fatigue. To adjust this deformation, it is recommended to use \textbf{carbon fiber cloth for reinforcement}... \\
\midrule
\textbf{Geo-Expert: Qwen3-8B-geo} \hfill \textbf{[Score: 9/10]} \\
\textit{Accurate Geological Reasoning:} \\
...Local thickening of a wedge leads to \textbf{internal stress concentration} and an increase in thrust... This deformation is primarily accommodated through \textbf{thrust fault sliding, extensional collapse, and lateral material flow}... \\
\bottomrule
\end{tabularx}
\end{table}

\section{Related Work}

\subsection{Large Language Models in Earth Sciences}
The application of AI in Earth sciences has grown rapidly, 
spanning medium-range global weather forecasting 
\citep{lam2023learning} and scientific language modeling 
\citep{taylor2022galactica}. Within geoscience specifically, 
two complementary lines of work have emerged. The first builds 
\emph{foundation models} via continued pre-training on broad 
geoscience corpora: K2 \citep{deng2024k2} adapts LLaMA on 
geoscience literature and pairs it with a curated knowledge 
graph, while GeoGalactica \citep{lin2023geogalactica} extends 
Galactica with over 65B geoscience-specific tokens. These 
models improve factual recall across general geoscience topics 
but do not explicitly target multi-step deductive reasoning. 
The second line develops \emph{geospatial agents and benchmarks}: 
GeoGPT \citep{zhang2023geogpt} demonstrates autonomous spatial 
query execution and facility siting from natural language, and 
UnivEARTH \citep{kao2025towards} evaluates LLM agents on Earth 
observation tasks using NASA satellite data. 

Despite this progress, a critical gap persists in 
\emph{solid-earth} geology. Foundation models like K2 and 
GeoGalactica are trained predominantly on surface-oriented 
literature (remote sensing, hydrology, atmospheric science), 
while agent-based systems operate on 2D geospatial workflows. 
Neither paradigm addresses the core challenges of structural 
geology and stratigraphy—cross-cutting relationship inference, 
polyphase tectonic restoration, and deep-time evolutionary 
reasoning grounded in abstract physical principles. 
\textsc{Geo-Expert} targets this neglected subdomain, shifting 
the adaptation objective from factual retrieval over surface 
phenomena to structured reasoning about subsurface geological 
processes.

\subsection{Domain-Specific Adaptation and PEFT}
General-purpose LLMs remain prone to hallucination in specialized 
STEM domains, motivating a growing body of domain-adapted models 
in medicine \citep{li2023chatdoctor}, finance 
\citep{wu2023bloomberggpt}, and law \citep{cui2023chatlaw}. 
A common limitation of these efforts is their emphasis on 
\emph{factual recall}—e.g., symptom lookup or terminology 
matching—rather than multi-step deductive reasoning. Recent 
work on reasoning-oriented distillation 
\citep{mukherjee2023orca,yu2023metamath} suggests that explicit 
chain-of-thought supervision is essential for eliciting logical 
inference, yet this insight has rarely been applied to 
domain-specific scientific adaptation.
Parameter-Efficient Fine-Tuning (PEFT), particularly Low-Rank 
Adaptation (LoRA) \citep{hu2021lora}, offers a practical path 
to internalize domain knowledge while preserving base-model 
capabilities. However, its effectiveness is fundamentally bounded 
by the quality of the supervision signal: tuning on raw domain 
text teaches the model \emph{what to say} but not \emph{how to 
reason}. In the geological setting this gap is especially acute, 
as correct answers often require chaining stratigraphic 
principles, cross-cutting relationships, and tectonic context 
across multiple inference steps. \textsc{Geo-Expert} addresses 
this by coupling LoRA with a CoT-enhanced instruction synthesis 
pipeline, shifting the adaptation objective from surface-form 
memorization to structured geological reasoning.

\subsection{Scientific Benchmarks and LLM-as-a-Judge}
Evaluating LLMs in specialized scientific domains poses a significant challenge. Traditional benchmarks often rely on static, multiple-choice questions extracted from standardized exams \citep{hendrycks2020measuring}. While useful for assessing factual recall, these static datasets fail to capture the multi-step deductive logic required in real-world scientific research. To overcome the limitations of human evaluation scalability, the ``LLM-as-a-Judge'' paradigm has been increasingly adopted to assess open-ended, complex reasoning tasks \citep{zheng2024judging}. Geo-Eval advances this paradigm by introducing an adversarial mining pipeline. By isolating ``hard boundary'' samples where general-purpose models fail but domain-aligned models succeed, Geo-Eval provides a much more rigorous and discriminative metric for assessing true expert-level geological reasoning.

\section{Conclusion and Future Work}
In this work, we present Geo-Expert, a framework that addresses the critical reasoning bottlenecks of generalist LLMs in highly specialized scientific domains. At the heart of our contribution is a robust, scalable dataset construction and instruction synthesis pipeline. Rather than passively fine-tuning, this pipeline achieves deep domain alignment by systematically recalibrating the model's semantic representations. By doing so, Geo-Expert successfully transforms text generation into well-rounded scientific reasoning.

Beyond achieving strong domain performance, this research provides evidence that parameter count alone is not sufficient in narrow expert domains. Our empirical results show that a domain-aligned 8B model can outperform a 70B open-weight generalist on Geo-Eval. This supports the practical value of high-quality data curation and targeted reasoning supervision for vertical scientific adaptation. We do not claim a new fine-tuning algorithm; rather, the contribution is an applied, reproducible recipe and benchmark for solid-earth geological reasoning.

Our work has several limitations. First, the training corpus is built from five canonical textbooks, biased toward structural geology, stratigraphy, and tectonics; mineralogy, geochemistry, and geophysics are under-represented. Second, the Geo-Eval benchmark contains 387 expert-vetted questions, sufficient for boundary-aware stress testing but smaller than general-purpose benchmarks and not large enough for fine-grained subcategory claims. Third, the benchmark construction and LLM-as-a-Judge scoring protocol may still contain selection and evaluator bias despite expert vetting. Fourth, the current comparison does not include direct runs of K2 or GeoGalactica on Geo-Eval, and we have not yet tested whether the same pipeline transfers to a second STEM discipline. Finally, the current framework is text-only and does not yet handle the inherently multimodal nature of geological data (cross-sections, well logs, field photos).

While Geo-Expert establishes a formidable foundation for text-based geological AI, we acknowledge that Earth sciences are inherently multimodal. Future work will extend this parameter-efficient framework by integrating Vision-Language Models (VLMs) to parse stratigraphic profiles and remote sensing imagery. Coupled with Retrieval-Augmented Generation (RAG) \citep{lewis2020retrieval} to dynamically track evolving literature, we aim to evolve Geo-Expert from a static reasoning engine into a dynamic, autonomous agent for geoscience discovery.

\section*{Availability and Responsible Release}
Code availability: the reproducibility code, including Markdown cleaning,
instruction-synthesis prompts, fine-tuning command templates, evaluation
scripts, and small self-authored examples, is available at
\url{https://github.com/Kunling-AI/Geo-Expert}. Because the instruction-tuning
corpus is derived from copyrighted textbooks, we do not redistribute raw
textbook text, the full textbook-derived training corpus, or long generated
passages that would reconstruct the sources. Model checkpoints and the full
benchmark set are not included in the public code repository at this stage.
The released scripts and examples are intended to support independent
reimplementation with licensed source material and local model paths.

\section*{Impact Statement}
This paper presents work whose goal is to advance the field of Machine Learning. There are many potential societal consequences of our work, none which we feel must be specifically highlighted here.

\section*{Acknowledgements}
We are grateful to the Editors and anonymous reviewers for their constructive
reviews and insightful suggestions, which significantly improve the manuscript.
This research was financially funded by National Natural Science Foundation of
China (No. 42202231, 42121005, 92141005). This work was supported by a grant
from the Postdoctoral Applied Research Foundation of Qingdao City. Furthermore,
we appreciate the editors and the reviewers for their critical, careful, and
constructive reviews. Figures in this paper were prepared by the authors based
on the Geo-Expert data-processing, instruction-synthesis, and fine-tuning
workflow.

\clearpage
\appendix
\bibliographystyle{icml2026}
\bibliography{references}
\onecolumn

\section{Implementation Details}
\label{app:implementation}

In this section, we provide the detailed hyperparameter configurations and hardware setups used for fine-tuning the Geo-Expert model family. All experiments were conducted using the \texttt{ms-swift} framework to ensure standardized PEFT across different model architectures. 

To balance training efficiency and memory footprint, we customized the LoRA settings based on the base models' parameter scales. For the compact Qwen3-8B model, we utilized a single NVIDIA RTX 5090 GPU with FP16 precision. For the larger architectures (Gemma-3-27B and Qwen3-32B), we scaled the hardware to a 4$\times$RTX 5090 and employed BF16 mixed-precision training. Gradient checkpointing was enabled in all runs. The detailed hyperparameter settings are summarized in Table \ref{tab:hyperparameters}.

\begin{table}[h]
\caption{Hyperparameter configurations for fine-tuning the Geo-Expert model family.}
\label{tab:hyperparameters}
\vskip 0.15in
\begin{center}
\begin{small}
\begin{sc}
\begin{tabular}{lcc}
\toprule
\textbf{Hyperparameter} & \textbf{8B Model} & \textbf{27B/32B Models} \\
\midrule
LoRA Target Modules & All Linear & All Linear \\
LoRA Rank ($r$) & 32 & 64 \\
LoRA Alpha ($\alpha$) & 32 & 128 \\
LoRA Dropout & 0.05 & 0.05 \\
Learning Rate & 2e-5 & 1e-4 \\
Optimizer & AdamW & AdamW \\
Precision & FP16 & BF16 \\
Gradient Checkpointing & True & True \\
Batch Size & 1 (per device) & 4 (per device) \\
Gradient Accumulation & 1 & 4 \\
Hardware & 1$\times$RTX 5090 & 4$\times$RTX 5090 \\
\bottomrule
\end{tabular}
\end{sc}
\end{small}
\end{center}
\vskip -0.1in
\end{table}

\section{Prompt Templates}
\label{app:prompts}

\subsection{LLM-as-a-Judge Prompt for Geo-Eval}
To conduct a rigorous and impartial evaluation during the hard sample mining phase of Geo-Eval, we utilized GLM-4.5 as the evaluator. We designed a zero-shot prompt template that constrains the judge to focus purely on geological factual accuracy and logical deduction, penalizing superficial or hallucinatory responses. The specific prompt used is as follows:

\begin{quote}
\textit{``You are an expert professor in solid earth geology. Your task is to evaluate the responses of two AI models to a complex geological question. \\
\textbf{Question:} [Insert Question] \\
\textbf{Model A Response:} [Insert Model A Answer] \\
\textbf{Model B Response:} [Insert Model B Answer] \\
\textbf{Evaluation Criteria:} 
1. Factual Accuracy (40\%): Does the response align strictly with established geological principles (e.g., Andersonian faulting, sequence stratigraphy)? 
2. Deductive Logic (40\%): Is the chain of thought coherent? 
3. Professional Terminology (20\%): Are the geological terms used accurately without cross-domain confusion? \\
\textbf{Task:} Score each model from 0 to 10. You must provide a brief justification for your score, explicitly pointing out any hallucinations or logical flaws. Output your final scores in the format: Score A: [X], Score B: [Y].''}
\end{quote}

\subsection{Reference-Based Evaluator Prompt for Final Scoring}
To conduct the final quantitative evaluation, we utilized GPT-4o as a reference-guided evaluator. Unlike zero-shot blind testing, this phase employs a strict Reference-Based Evaluation paradigm, where the model compares candidate responses against expert-verified ground truth. The prompt explicitly enforces terminology flexibility to accurately accommodate the nuances of geological nomenclature.

\vspace{0.1in}
\begin{verbatim}
You are an authoritative Professor of Geology and an expert academic evaluator. 
Your task is to grade a Model's Answer to a complex geological question against 
a provided professional Reference Answer.

[Grading Criteria]
1. Score from 0 to 10.
2. Focus on SEMANTIC CORRECTNESS and GEOLOGICAL ACCURACY, not verbatim matching.
3. Terminology Flexibility: Be lenient with accepted geological synonyms
   (e.g., "igneous" vs "magmatic", "thrust fault" vs "reverse fault"
   depending on context). 
4. Completeness: A 10/10 answer must contain all core geological facts mentioned 
   in the reference.
5. Penalty: Deduct heavily for hallucinations, factually incorrect statements, 
   or mixing up crucial concepts (e.g., misidentifying rock types or eras).

[Output Format]
Output ONLY a JSON object with this exact structure: 
{"score": 8, "reason": "brief professional explanation"}

Question: {question}
Reference Answer: {reference}
Model Answer: {model_ans}
\end{verbatim}
\vspace{0.1in}

\subsection{Instruction Synthesis Details}
The instruction synthesis stage uses DeepSeek-R1 as the primary reasoning-oriented teacher model. For reproducibility, we release a command-line implementation that performs Markdown chunking, source-grounded question generation, CoT answer construction, filtering, and ShareGPT-style export. For each cleaned chunk, the prompt contains the local Markdown content, chapter path, domain tags from the domain tree, and an explicit requirement that the answer remain grounded in the provided source chunk. We use temperature 0.6 and a maximum generation length of 2048 tokens for candidate generation. Post-processing removes invalid chat schemas, duplicate questions, empty answers, and responses that fail simple source-grounding checks such as introducing unrelated disciplines or unsupported entities. The resulting records are stored in a chat format with \texttt{user} and \texttt{assistant} messages. When a reasoning trace is generated, it is wrapped in \texttt{<think>} tags before the final answer.

\subsection{Instruction-Tuning Data Example}
As described in Section 2.2, our instruction synthesis pipeline generates high-quality, reasoning-enhanced QA pairs. Below is a translated JSON snippet extracted from our training corpus, demonstrating the required \texttt{<think>} trajectory before final generation.

\vspace{0.1in}
\begin{verbatim}
[
  {
    "messages": [
      {
        "role": "user",
        "content": "How is the method of measuring vorticity using rotated 
                    porphyroblasts applied to the study of rocks near the 
                    Alpine Fault?"
      },
      {
        "role": "assistant",
        "content": "<think>\n1. Identify core concepts: vorticity measurement, 
                    rotated porphyroblasts, Alpine Fault.\n2. Recall geological 
                    context: The Alpine Fault is a major plate boundary fault 
                    with intense ductile deformation.\n3. Formulate explanation: 
                    Porphyroblasts record kinematic history during shear 
                    deformation...\n</think>\n\nThe method of measuring 
                    vorticity using rotated porphyroblasts is a sensitive 
                    kinemetric technique used to study the behavior of rocks 
                    during shear deformation. By analyzing the rotational 
                    dynamics of these crystals within the ductile shear zones..."
      }
    ]
  }
]
\end{verbatim}
\vspace{0.1in}

\subsection{Geo-Eval Hard Boundary Sample Analysis}
In Section 2.4, we detailed the automated mining of ``hard boundary'' samples where general-purpose models fail but domain-expert models succeed. The following real evaluation cases illustrate severe cross-domain hallucinations by the base model and precise domain anchoring by Geo-Expert.

\vspace{0.1in}
\noindent \textbf{Case 1: Semantic Disambiguation of Polyphase Deformation} \\
\textbf{Question:} What is polyphase deformation? \\
\textbf{Ground Truth:} In structural geology, Polyphase Deformation (or Superimposed Deformation) refers to the complex process where the same geological body undergoes two or more distinct regional tectonic stress fields over a long period of geological history...

\vspace{0.05in}
\noindent \textbf{\textit{Base Model Response (Score: 0.0 / 10):}} \\
\textbf{[Exposed Thinking Process]:} \textit{Hmm, the user is asking about ``polyphase deformation''. First, I need to determine the context of this term. It could appear in materials science, engineering, or computer graphics. For instance, in computer graphics, it might involve multiple steps of deformation algorithms...} \\
\textbf{[Final Output]:} \textit{``Polyphase deformation'' is an interdisciplinary concept. In materials science, it refers to different stages of deformation under varying temperatures or stress... In computer graphics, it refers to the combination of multiple animation steps...}

\vspace{0.05in}
\noindent \textbf{\textit{Geo-Expert Response (Score: 9.0 / 10):}} \\
\textbf{[Final Output]:} \textit{In structural geology, Polyphase Deformation refers to the complex superimposed deformation process generated by two or more distinct tectonic stress fields over geological time. Each independent deformation event leaves specific structural traces in the rock, such as folds, faults, or foliations. Studying polyphase deformation is of decisive significance for reconstructing paleostress fields and understanding tectonic evolution...}

\vspace{0.15in}
\noindent \textbf{Case 2: Disciplinary Anchoring of ``Deformation'' Terminology} \\
\textbf{Question:} What is the academic definition of deformation terminology? \\
\textbf{Ground Truth:} The academic definition of deformation refers to the process and result of relative positional changes among internal particles of an object under external factors (e.g., stress, temperature), leading to changes in shape, size, orientation, or volume. It is a core concept in structural geology and solid mechanics...

\vspace{0.05in}
\noindent \textbf{\textit{Base Model Response (Score: 0.0 / 10):}} \\
\textbf{[Exposed Thinking Process]:} \textit{Hmm, the user is asking about the academic definition of ``deformation terminology''. This could have different interpretations in linguistics, philosophy, or sociology. For instance, in linguistics, it might refer to inflection or word morphing. Or it could mean a term that is morphed or transformed across contexts...} \\
\textbf{[Final Output]:} \textit{``Deformation terminology'' does not have a universally accepted definition. Its meaning varies by discipline: In linguistics, it may refer to word inflection or semantic shifts. In philosophy, it may refer to conceptual evolution. In sociology, it implies the contextualization of a term across different social groups...}

\vspace{0.05in}
\noindent \textbf{\textit{Geo-Expert Response (Score: 9.0 / 10):}} \\
\textbf{[Final Output]:} \textit{The academic definition of deformation terminology refers to the scientific concepts used to describe the shape changes of rocks under stress, encompassing major types such as elastic, plastic, and brittle deformation. Elastic deformation refers to reversible shape changes... Furthermore, it includes key concepts like stress, strain, and strain rate, as well as deformation mechanisms (e.g., dislocation creep). These terms collectively form a systematic description of rock deformation processes, providing a theoretical foundation for studying geological structures.}

\end{document}